\documentclass{article} 
\usepackage{colm2024_conference}
\usepackage[normalem]{ulem}
\useunder{\uline}{\ul}{}
\usepackage{microtype}
\usepackage{hyperref}
\usepackage{url}
\usepackage{booktabs}
\usepackage[notransparent]{svg}
\usepackage{amsmath}
\usepackage{graphicx}
\usepackage{comment}

\title{Optimising Calls to Large Language Models with Uncertainty-Based Two-Tier Selection}

\author{Guillem Ramírez \\
University of Edinburgh\\
\texttt{gramirez@ed.ac.uk} \\
\And
Alexandra Birch\\
University of Edinburgh\\
\texttt{A.Birch@ed.ac.uk} \\
\And
Ivan Titov\\
University of Edinburgh\\
University of Amsterdam\\
\texttt{ititov@inf.ed.ac.uk} \\
}

\colmfinalcopy 
\begin{document}

\maketitle

\begin{abstract}
Researchers and practitioners operating on a limited budget face the  cost-performance trade-off dilemma. The challenging decision often centers on whether to use a large LLM with better performance or a smaller one with reduced costs. This has motivated recent research in the optimisation of LLM calls. Either a \textit{cascading} strategy is used, where a smaller LLM or both are called sequentially, or a \textit{routing} strategy is used, where only one model is ever called. Both scenarios are dependent on a decision criterion which is typically implemented by an extra neural model. In this work, we propose a simpler solution; we use only the uncertainty of the generations of the small LLM as the decision criterion. We compare our approach with both cascading and routing strategies using three different pairs of pre-trained small and large LLMs, on nine different tasks and against approaches that require an additional neural model. Our experiments reveal this simple solution optimally balances cost and performance, outperforming existing methods on 25 out of 27 experimental setups.
\end{abstract}

\pdfoutput=1


\author{First Author \\
  Affiliation / Address line 1 \\
  Affiliation / Address line 2 \\
  Affiliation / Address line 3 \\
  \texttt{email@domain} \\\And
  Second Author \\
  Affiliation / Address line 1 \\
  Affiliation / Address line 2 \\
  Affiliation / Address line 3 \\
  \texttt{email@domain} \\}

\section{Introduction}
Large Language Models (LLMs) offer a high performance for a wide range of text tasks. Their widespread popularity both in research and industrial applications necessitates an understanding on how to optimally use them. Bigger models tend to have better performance, while smaller models are faster and cheaper to run. Deciding which model to use is a common dilemma for many researchers and practitioners with limited budgets, time-constraints or environmental concerns.  

\begin{figure}
  \centering
\includegraphics[trim={3.5cm 8.5cm 1.8cm 11cm},clip, width=0.7\linewidth]{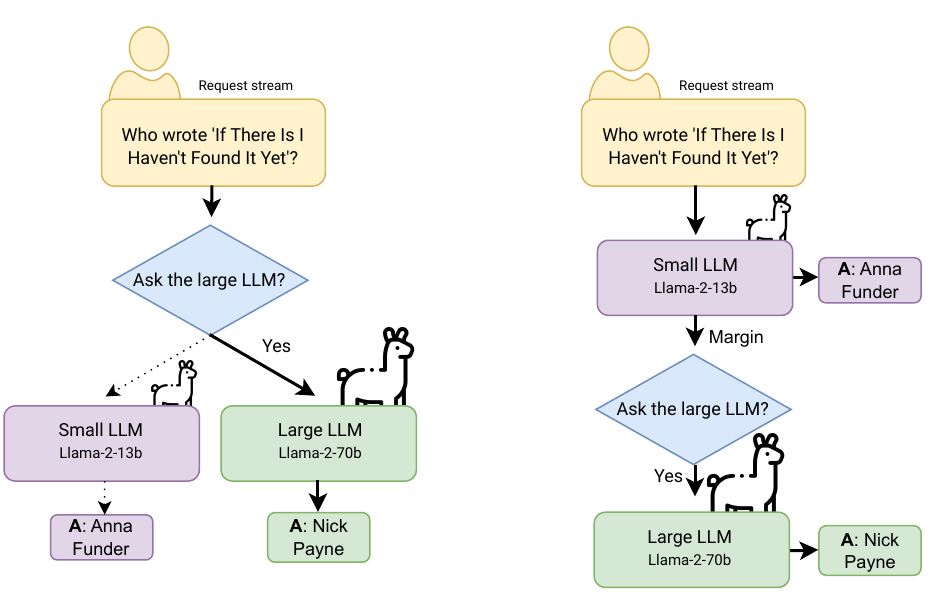}
    \caption{Routing (left) attempts to select the LLM with the best cost-accuracy trade-off given an incoming query. In cascading (right), all queries are passed through the small model, and depending on its output, the large LLM is consulted. We propose using a cascading approach that uses the margin of the generations to score outputs from the small LLM.}
    \label{diagram}
\end{figure}

Recent works attempt to optimise calls to a set of LLMs. In this work we consider the set-up with two LLMs, where one is more expensive with greater performance than the other. In this scenario, there are two main strategies (Figure~\ref{diagram}): \textit{routing}, where a query from a user is directed to only one model based on a decision criterion; \textit{cascading}, where the query always goes to the cheaper model and may subsequently go to the more expensive model depending on the cheaper model's output. These previous studies use one of these calling strategies, and involve either using an auxiliary model to score an LLM output ~\citep{frugalgpt, sakota, hybrid, automix} or using repeated calls to the small cheaper LLM ~\citep{qbc, automix}.

Studies that use an auxiliary model introduce further complexity in the optimisation approach, and it remains unclear when practitioners should rely on these auxiliary models. Not only do they require additional training, but they also usually require specific training data and the auxiliary models may not generalise to other tasks. For studies that rely on repeated calls to the small LLM, this approach can become expensive, undermining the original practical motivation for its use. It is perhaps for these reasons, that neither approach has gained traction among practitioners.

However, we question whether these additional models or the repeated calls are required to optimise LLM calls. We hypothesise that they may be unnecessary in many use cases, as we can extract confidence measures from the generations of the small model. To investigate this, we propose a \textit{cascading} policy that uses a simple measure of confidence of the small LLM to decide whether the large LLM needs to be called. We evaluate our policy on nine tasks, and for three different pairs of pre-trained small and large LLMs. We focus specifically on short-generation tasks, which are widely used and versatile, and do not include long-generation tasks in this study due to the additional complexity they introduce for evaluation. In our experiments on classification, multiple-choice and Question-Answering (QA) tasks, this policy outperforms methods that require additional training. 

We believe our findings suggest that the optimisation of LLM calls should gravitate towards understanding their existing available signals, as opposed to training and running further auxiliary models. Moreover, simple and cheap policies have a lower entry cost for researchers and practitioners, which has an impact on their adoption.

The key contributions of this work are as follows:
\begin{itemize}
    \item We propose using the margin of the generations of LLMs to optimise LLM calls. The approach is simple, non-parametric, and does not require data or running multiple calls to the small model.
    \item We test our policy on nine tasks with three different pairs of small and large LLMs, and against relevant neural methods that use auxiliary models. We find that Margin Sampling outperforms the other methods, despite using fewer resources. In addition, we further test our policy in a multi-task set-up and obtain similar results. 
    \item Our results underscore the importance of understanding signals from LLMs in the context of optimising calls to them, in contrast with previous work, which require additional training of auxiliary models.  
\end{itemize}

\section{Related work}

\paragraph{LLM uncertainty} The margin between the two most likely classes has been widely regarded as an uncertainty measure adopted from early Active Learning literature \citep{scheffer_active_2001, luo_active_2004} and considered more recently in the context of Knowledge Distillation from LLMs~\citep{baykal_robust_2023,cache-distil}. Other measures of uncertainty exist for LLMs~\citep{uncertainty2, uncertainty}. The primary challenge in text generation lies in the difficulty of distinguishing between various forms of uncertainty. Specifically, when evaluating the confidence level of a text generator, we would ideally want to concentrate on uncertainties related to the change in meaning. This entails distinguishing uncertainties that affect the intended meaning from meaning-preserving variations in the generated text. In this study, we address this challenge by concentrating on tasks that require generating very short sequences of tokens. We find that even a basic approach to uncertainty can be advantageous. Additionally, while it might seem logical to assess uncertainty through multiple text outputs from a model, this would lead to significant computational costs, which are impractical for our purposes. Therefore, we employ a straightforward method that does not necessitate generating multiple text samples.

\paragraph{Optimisation of inference costs} Several methods have been proposed to improve the latency of LLMs, such as speculative decoding~\citep{speculative}, knowledge distillation~\citep{KD1,KD2} and model quantisation~\citep{quantisation}. However, operational costs rather than latency are the focus of this work. Our method only requires access to the logits of a small pre-trained LLM with no need for its parameters. 

\paragraph{Optimisation of LLM API Calls} Recent work deals with the problem of optimising calls to a pool of LLMs~\citep{survey}. \citet{sakota} and \citet{routingexpert} propose to train an auxiliary model that predicts the success of calling each LLM. 
Similarly, \citet{hybrid} developed a model that predicts the quantitative benefit of utilizing a smaller language model over a larger one.

\citet{frugalgpt} proposed \textit{cascading}: using an auxiliary model to predict the accuracy of the small LLM's output. \citet{automix} and \citet{ecoassistant} used cascading in conjunction with multiple calls to the small model. Finally, \citet{qbc} propose a cascading approach that requires repeated calls of the small LLM for reasoning tasks; we show that this can be simplified to just looking at the most likely tokens for short-generation tasks. 

\citet{cache-distil} showed that the margin on a knowledge-distilled model could optimise calls to the larger LLM. This work was limited to one pre-trained large LLM and a smaller fine-tuned local model. However, many practitioners and researchers may not be able to fine-tune their own models due to budget constraints. We extend these findings to pairs of both small and large pre-trained LLMs, as well as to other generation tasks including a multi-task setup. 

Our proposed method differs from previous studies as it does not require previously annotated data for the task and does not require repeated calls to the small LLM. Finally, it does not require training and deploying an auxiliary model.

\section{Optimisation of LLM calls}

\subsection{Problem definition}

In this work, we predict mappings between elements in the input space, $\mathcal{X}$, and the corresponding labels in the output space, $\mathcal{Y}$. We have access to the small and the large LLMs that we can prompt to become predictors $f_{\text{s}}, f_{\text{l}}: \mathcal{X} \rightarrow \Delta(\mathcal{Y})$, where $\Delta(\mathcal{Y})$ denotes the class of probability distributions over $\mathcal{Y}$. We simulate the online setting, where users send queries sequentially. We have $q$ queries ($x_1, \dots, x_q$) $\overset{\mathrm{iid}}{\sim}$ $\mathcal{X}$ and we predict the corresponding labels ($y_1, \dots, y_q$). For each incoming query $x_i$, we decide whether to call an LLM based on a calling strategy (see below), and incur a given cost $c_{\text{s}}(x_i)$ or $c_{\text{l}}(x_i)$ respectively. The average cost of the queries by both models is then given by $\hat{c}_{\text{s}} = \frac{1}{q} \sum_{i} c_{\text{s}}(x_i)$; $\hat{c}_{\text{l}} = \frac{1}{q} \sum_{i} c_{\text{l}}(x_i)$. We assume that $\hat{c}_{\text{s}} < \hat{c}_{\text{l}}$.

\subsection{LLM Calling Strategies}
For strategies that require training an additional model, we use a train dataset $X_{\text{train}}, Y_{\text{train}}$, and use either the corresponding labels from the small LLM only, $f_{\text{s}}(X_{\text{train}})$, or from both LLMs, $f_{\text{s}}(X_{\text{train}})$ and $f_{\text{l}}(X_{\text{train}})$, depending on the strategy. For the training of the auxiliary models, we follow the original papers as much as possible and perform a hyperparameter search where values are omitted. See Appendix \ref{implement} for a detailed explanation of the training process. 

\subsubsection{Routing Strategies}
Since only the small or the large LLM is called in routing strategies, then for a given target average cost per query $c$ ($\hat{c}_{\text{s}} \leq c\leq \hat{c}_{\text{l}} $), we call the large LLM with a probability $p_{\text{r}}$ calculated from the re-arranged form of Equation~\ref{random}.
\begin{equation}
    c = (1-p_{\text{r}})\hat{c}_{\text{s}} + p_{\text{r}} \hat{c}_{\text{l}}
    \label{random}
\end{equation}

\paragraph{Random routing} For every incoming query we call the large LLM with the probability $p_{\text{r}}$.

\paragraph{Routing~\citep{sakota,routingexpert}} We train a meta-model to predict the performance of the small LLM only, given an incoming query. If this prediction is below a threshold value related to the probability $p_{\text{r}}$, it indicates the small LLM's performance is insufficient and thus we must call the large LLM. 

\paragraph{HybridLLM~\citep{hybrid}} The performance of both the small and large LLMs are modelled in this strategy. We train a meta-model to predict if an incoming query is likely to be better solved by the small LLM than by the large LLM. As in Routing above, if this prediction is below a threshold value related to the probability $p_{\text{r}}$, we call the large LLM, otherwise the small LLM. 

\subsubsection{Cascading}

Since the small LLM is always called, then for a given target average cost per query, $c$, we call the large LLM with a probability $p_{\text{c}}$ calculated from the re-arranged form of Equation~\ref{cascading_cost}.
\begin{equation}
    c= \hat{c}_{\text{s}} + p_{\text{c}} \hat{c}_{\text{l}}
    \label{cascading_cost}
\end{equation}

\paragraph{FrugalGPT~\citep{frugalgpt}} We train a model that, given a query and a candidate answer, predicts if the latter is correct. If this prediction is below a threshold value related to the probability $p_{\text{c}}$, we call the large LLM. 

\paragraph{Margin Sampling (ours)} We suggest using the uncertainty of the output, namely the margin~\citep{scheffer_active_2001, luo_active_2004}, defined by:
\begin{equation}
\begin{aligned}
 \text{Margin}_{f_{\text{s}}}(x_i) = 
    P_{f_{\text{s}}}(y_i = k_1^{t = 1} \mid x_i) - P_{f_{\text{s}}}(y_i = k_2^{t = 1} \mid x_i)
\end{aligned}
\end{equation}
where $k_1^{t = 1}$ and $k_2^{t = 1}$ are the first and second most likely tokens, respectively, according to the distribution of $f_{\text{s}}$ for the first predicted token position, $t = 1$. One advantage of this approach is that it does not require generating a full sequence to be able to compute uncertainty. Moreover, for the tasks we consider there is generally more uncertainty in the first token. If the margin is below a threshold value related to the probability $p_{\text{c}}$, we call the large LLM.

\subsection{Dynamic threshold}

All the investigated strategies require setting a threshold for the decision criterion, and we select a dynamic threshold in this work. An initial threshold is calculated using the first 10 queries. We do not evaluate whether to call the large LLM for these 10 queries, we only obtain outputs from the auxiliary models, or the margin value for Margin Sampling. This may require calling the small LLM depending on the strategy. We then use this distribution to calculate an initial $p_{\text{r}}$ or $p_{\text{c}}$-th percentile value. For all subsequent queries, the decision to call the large LLM is made, and the threshold is dynamically updated based on all past queries. 
\section{Experimental setup}
\subsection{LLMs}
We study three pairs of small and large LLMs in our experiments: Mistral 7B~\citep{mistral} and Mixtral 8x7B \citep{mixtral}; Llama-2 of size 13B and Llama-2 of size 70B ~\citep{llama2}; GPT-3~\citep{gpt-3} and GPT-4~\citep{gpt-4}. We selected these pre-trained LLMs because of their popularity among practitioners, as well as to show robustness across different scales. We have arranged the pairs this way to keep the family of LLMs similar; however, this is not a requirement for any of the calling strategies. 

For the open-source families (Mistral, Llama-2), all our experiments are done locally in one NVIDIA A100 GPU (80 GB), after applying a 4-bit quantisation. Section~\ref{llms_used} contains more details about the LLMs used.

\subsection{Datasets}
We draw inspiration from \citet{helm} and \citet{cache-distil} 
and choose a wide range of tasks, showcasing different difficulties. The sizes of the datasets, along with the label distribution and accuracy of the LLMs, can be found in Appendix (Section~\ref{appendix:datasets}, Section~\ref{llms_used}).
\paragraph{Classification tasks} For emotion classification we use ISEAR~\citep{shao_universality_2015}; for fact-checking we use FEVER~\citep{thorne_fever_2018}; for sentiment analysis we use RT-Polarity ~\citep{pang_seeing_2005}, CR~\citep{cr}, and SST-2~\citep{sst2}. All of these datasets are balanced. 

\paragraph{Multiple-choice} We use Openbook~\citep{openbook}, a popular multiple-choice dataset that involves common knowledge of the world. 
\paragraph{QA - short generation} We use NaturalQuestions~\citep{DBLP:journals/tacl/KwiatkowskiPRCP19}, that contains real questions from human users; we use Wikifact~\citep{wikifact2, wikifact}, which consists of a knowledge base completion problem to test factual knowledge; we use bAbI~\citep{babi}, that tests language understanding and reasoning.

\subsection{Experiment details}
For each dataset we set aside $1,000$ data-points for a train dataset that is used only to train the auxiliary models. The remainder is used for the online test set. We use $n=500$ data-points from the train dataset in the Routing, HybridLLM and FrugalGPT strategies when training the auxiliary model, unless stated otherwise. For the auxiliary models, we fine-tune DistilBERT~\citep{distilbert}, as per \citet{sakota, hybrid,frugalgpt,automix,hybrid}. We further split the training data into 80\% train and 20\% validation, and fine-tune for 100 epochs with early stopping and patience of 20 epochs.

Unless stated otherwise, our results assume a simple cost scheme with $c_{\text{s}}(x_i)=1$ and $c_{\text{l}}(x_i)=10$, consistent with the pricing of commercial APIs~\citep{survey} and similar to the cost schemes of related work~\citep{automix, frugalgpt, routingexpert, sakota}. We do not take into account in our experiments the latency of running DistilBERT, which we deem negligible compared to running the LLMs. To evaluate accuracy across budgets, we report Area Under the Curve (AUC) of the accuracy divided by $ \hat{c}_{\text{l}}-\hat{c}_{\text{s}}$. Bolded results mark best performance, and underlined results mark second-best. We run our experiments with three random seeds, and report average results. 

\begin{figure*}[ht] 
    \centering
{\includegraphics[height=140pt]{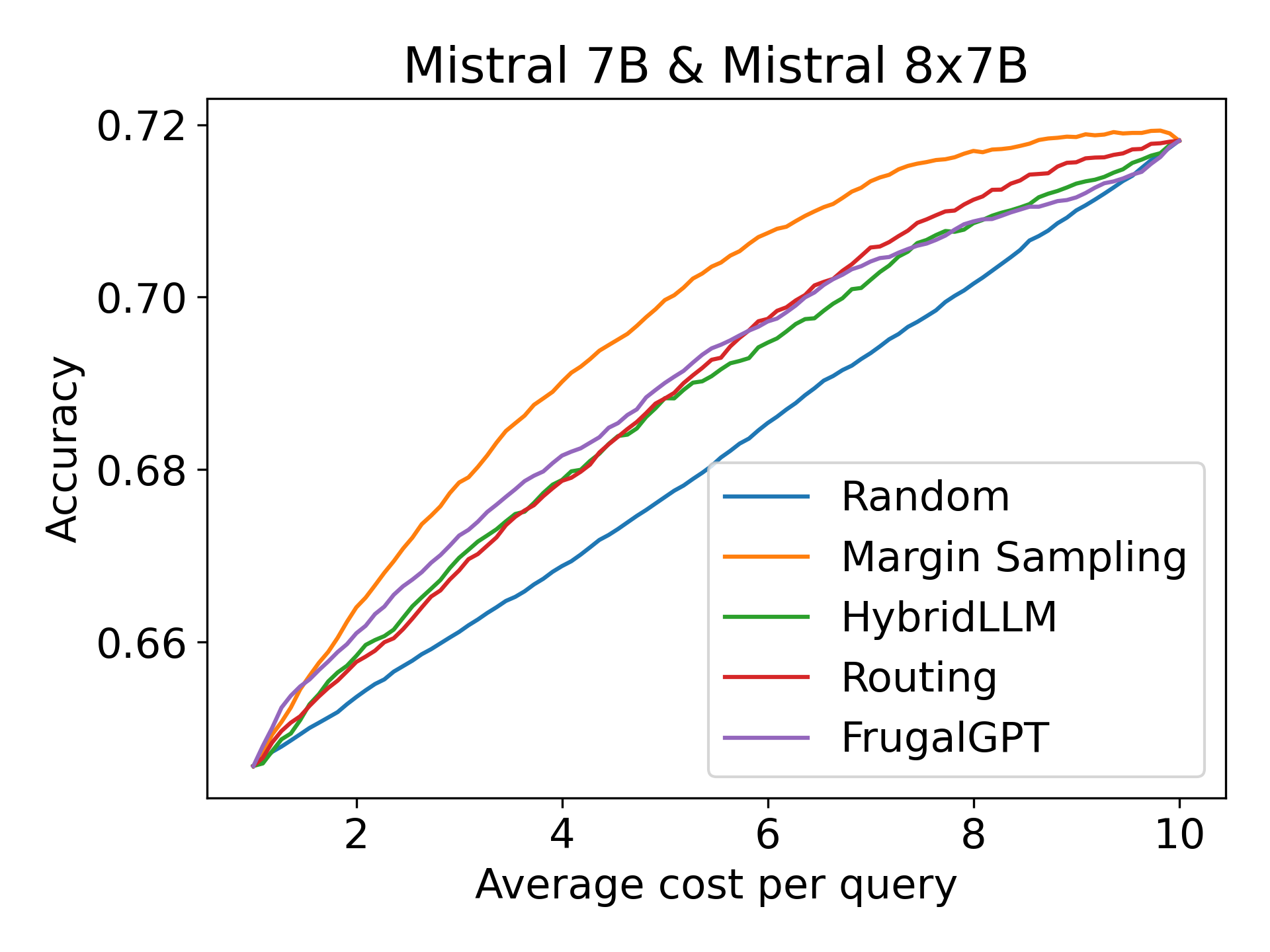}}
{\includegraphics[height=140pt]{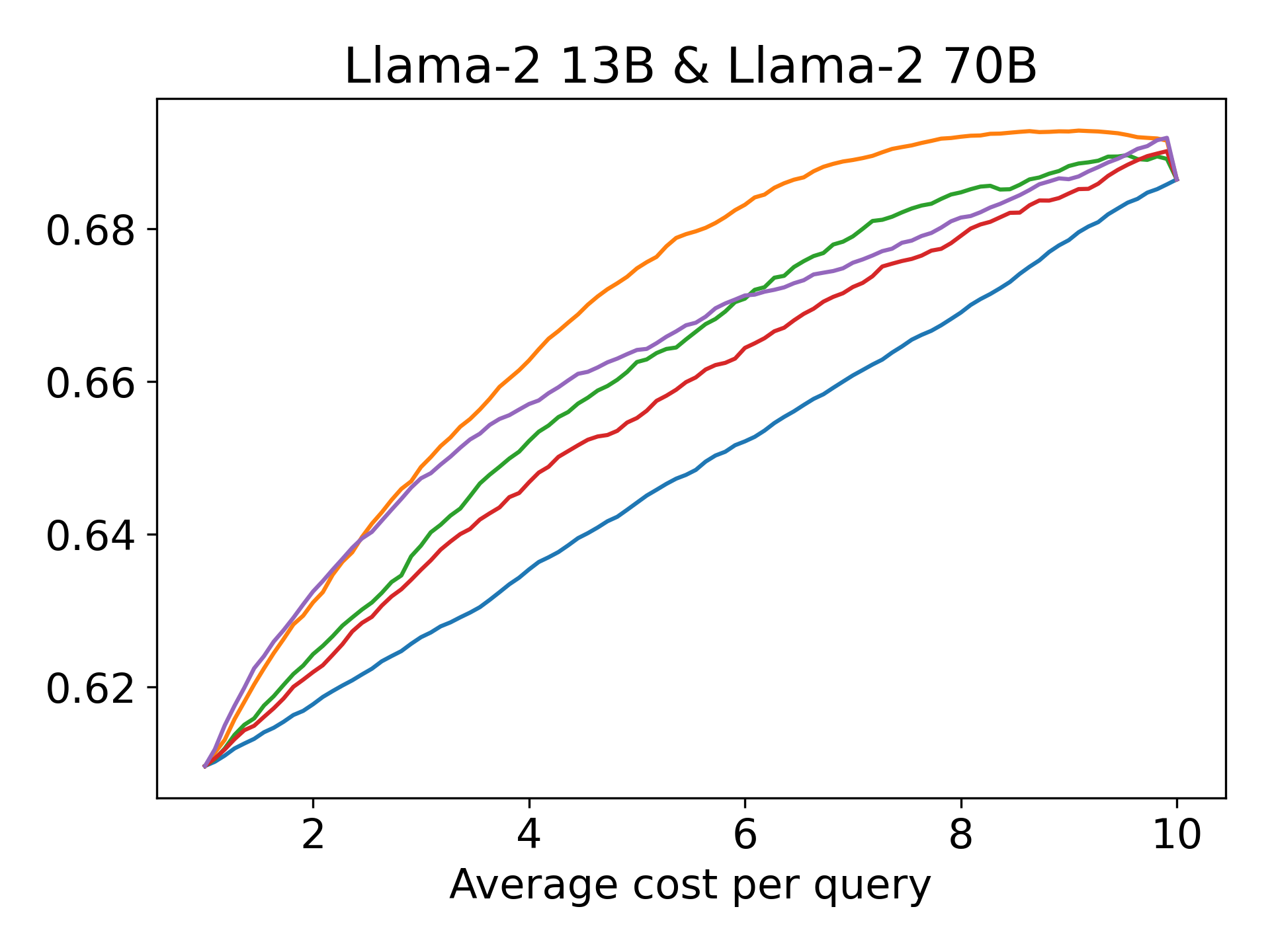}}
{\includegraphics[height=140pt]{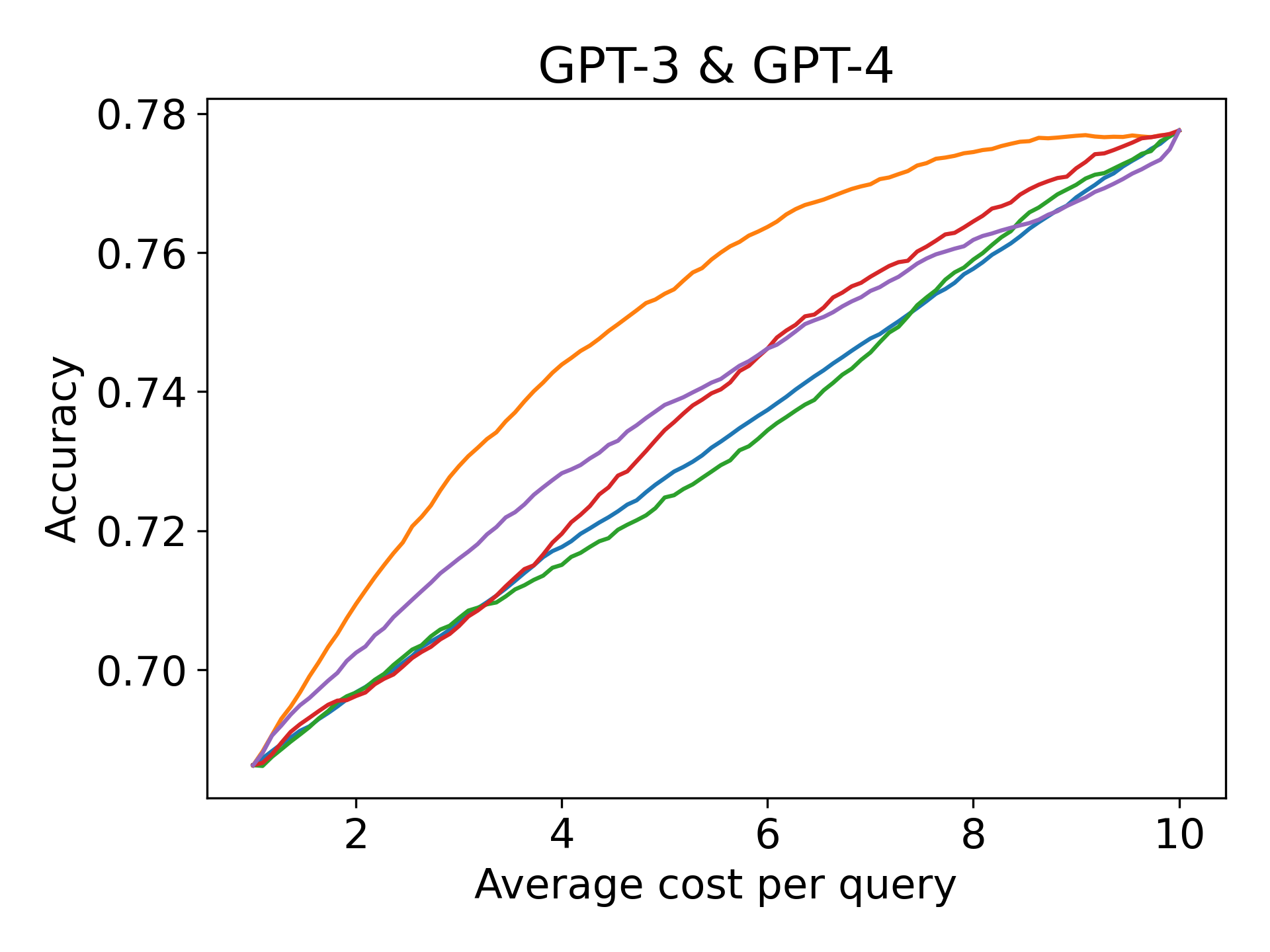}} 

    \caption{Accuracy curve with respect to budgets. We have averaged results for all the tasks.}
    \label{plots}
\end{figure*}

\section{Results}

\subsection{Comparison of calling strategies}
\label{mainresults}

Table ~\ref{main_results} shows the AUC for all five evaluated calling strategies, across the nine tasks and for the three pairs of LLMs. Averaged across all nine tasks (the final column), we see that Margin Sampling outperforms all strategies for all LLM pairs. Across individual tasks, we see that it performs best or second best for seven of the nine tasks consistently across all three LLM pairs. Of all the nine tasks and three LLM pairs, it does not achieve best or second-best performance for only two of the 27 combinations: for ISEAR and for NaturalQuestions for the Mistral 7B and Mixtral 8x7B LLM pair. This is likely due to the poor accuracy of the small LLM Mistral 7B on these tasks (Appendix~\ref{llms_used}). Figure~\ref{plots} shows how performance changes with budget. We see that Margin Sampling dominates across all budgets, despite not using any training data.  

Indeed, the performance of Margin Sampling seems to improve as the performance of the cheaper LLM improves, which is to be expected; it achieves its best results when applied on top of GPT-3. 

FrugalGPT is on average the second-best performing strategy. This is due to its good performance on classification tasks, which is expected as it uses a classifier trained on the task as the auxiliary model. However, it performs worse than the random baseline on the more challenging multiple-choice task, Openbook. FrugalGPT also performs inconsistently for QA tasks; we conclude that FrugalGPT may be satisfactory on relatively easy classification tasks and struggle with harder generation tasks. 

Finally, Routing and HybridLLM seem to have a good performance in QA tasks while having a worse performance in classification tasks. We note that HybridLLM on average has the same performance as random for the OpenAI models, which is a surprising finding.  

We have not shown comparisons to approaches that require repeated calls to a small LLM throughout this work, as in preliminary experiments we found they do not perform well (see Section~\ref{multiple_calls}). 

\addtolength{\tabcolsep}{-0.47em}
\begin{table*}[t]
\begin{tabular}{lcccccccccc}
\toprule
\multicolumn{1}{c}{} & \footnotesize \textbf{ISEAR} & \footnotesize \textbf{RT-Pol} & \footnotesize \textbf{FEVER} & \footnotesize \textbf{CR} & \footnotesize \textbf{SST-2} & \footnotesize \textbf{Openbook} & \footnotesize \textbf{Wikifact} & \footnotesize \textbf{bAbI} & \footnotesize \textbf{NaturalQ} & \footnotesize \textbf{Average} \\
\midrule
\multicolumn{11}{c}{Mistral 7B - Mixtral 8x7B} \\
\small Random               & \small 0.606                & \small 0.876                & \small 0.773                & \small 0.923                & \small 0.880                & \small 0.843                & \small 0.443                & \small 0.597                & \small {\ul 0.193}          & \small 0.681                \\

\small  Routing              & \small  {\ul 0.618}          & \small  0.876                & \small  \textbf{0.777}       & \small  0.924                & \small  0.890                & \small  0.844                & \small  {\ul 0.492}          & \small  {\ul 0.606}          & \small  0.177                & \small  0.689\\
\small  HybridLLM            & 
\small  {\ul 0.618}          & \small  0.876                & \small 0.776                & \small 0.924                & \small 0.886                & \small {\ul 0.849}          & \small 0.454                & \small \textbf{0.612}       & \small \textbf{0.199}       & \small 0.688                \\
\small  FrugalGPT            & \small \textbf{0.632}       & \small \textbf{0.887}       & \small \textbf{0.777}       & \small {\ul 0.931}          & \small \textbf{0.901}       & \small 0.835                & \small 0.477                & \small 0.596                & \small 0.172                & \small {\ul 0.690}          \\
\small  Margin Sampling      & \small  0.617                & \small {\ul 0.885}          & \small \textbf{0.777}       & \small \textbf{0.933}       & \small {\ul 0.899}          & \small \textbf{0.868}       & \small \textbf{0.499}       & \small {\ul 0.606}          & \small 0.187                & \small \textbf{0.697}  \\  
\midrule
\multicolumn{11}{c}{Llama-2 13B - Llama-2 70B} \\
\small Random               & \small 0.630                & \small  0.809                & \small  0.653                & \small  0.885                &  \small  0.873                &  \small 0.617                &  \small 0.505                &  \small 0.600                &  \small 0.259        & \small  0.648                 \\
\small Routing              &  \small 0.639          & \small 0.836                & \small 0.662      & \small 0.909                & \small 0.883                & \small 0.621                & \small {\ul 0.514} & \small 0.593          & \small 0.254                & \small 0.657                  \\
\small HybridLLM            & \small  0.641          & \small 0.844                & \small {\ul 0.681}          & \small 0.899                & \small 0.874                & \small {\ul 0.626}          & \small {\ul 0.514}       & \small {\ul 0.608} & \small {\ul 0.264} & \small 0.661                          \\
\small FrugalGPT            & \small \textbf{0.662}       & \small  \textbf{0.856}       & \small  0.668    & \small 
 \textbf{0.918} & \small  \textbf{0.899}       & \small  0.598                & \small 
 0.507                & \small  0.602                & \small  0.258                & \small 
 {\ul 0.663}   \\
\small Margin Sampling      & \small {\ul 0.645}    & \small {\ul 0.853}          & \small \textbf{0.691} & \small {\ul 0.912} & \small {\ul 0.893}    & \small \textbf{0.640}      & \small \textbf{0.516}            & \small \textbf{0.609} & \small \textbf{0.270}       & \small \textbf{0.670}  \\
\midrule
\multicolumn{11}{c}{GPT-3 - GPT-4} \\
\small Random               & \small 0.747	& \small 0.914 & \small 	0.816	& \small  0.931	& \small 0.898	& \small  0.877	& \small  0.552	& \small  0.574 & \small 0.281 & \small 	0.732    \\
\small  Routing              & \small  { \ul 0.769} & \small 	0.915	& \small  0.815	& \small 
 {\ul 0.936}	& \small  0.898	& \small  0.878	& \small 0.558	& \small  {\ul 0.584}	& \small  0.277 & \small 0.737 \\
\small  HybridLLM            &  \small 0.744	 &  \small 0.914	 &  \small {\ul 0.821} &  \small	0.932	 &  \small {\ul 0.899}	 &  \small {\ul 0.882} &  \small	0.556	 &  \small 0.558	 &  \small 0.278 &  \small	0.732  \\
\small  FrugalGPT            & \small 0.767	 & \small {\ul 0.922} & \small	0.819 & \small	\textbf{0.940}	 & \small \textbf{0.903} & \small	0.876	 & \small {\ul 0.564}	 & \small 0.572	 & \small {\ul 0.283} & \small	{\ul 0.738} \\
\small  Margin Sampling      & \small \textbf{0.771}	& \small \textbf{0.925}	& \small \textbf{0.826}	& \small \textbf{0.940}	& \small {\ul 0.899}	& \small \textbf{0.918}	& \small \textbf{0.584}	& \small \textbf{0.598}	& \small \textbf{0.294}	& \small \textbf{0.751}  \\
\bottomrule 
\end{tabular}
\caption{Accuracy (AUC) for the three LLM model pairs across the five classification tasks (columns 2-6), the multiple-choice task (column 7) and the three generation tasks (columns 8-10).}
\label{main_results}
\end{table*}

\subsection{Multi-task setting}
LLMs are often used to handle various tasks simultaneously. To simulate this scenario, we create an artificial multi-task setting by merging the datasets from all nine tasks. We then sample 10,000 data-points. We split this dataset into a 10$\%$ train set ($n=1,000$ data-points) and a 90$\%$ online test set. 

\begin{figure*}[t] 
    \centering
{\includegraphics[height=140pt]{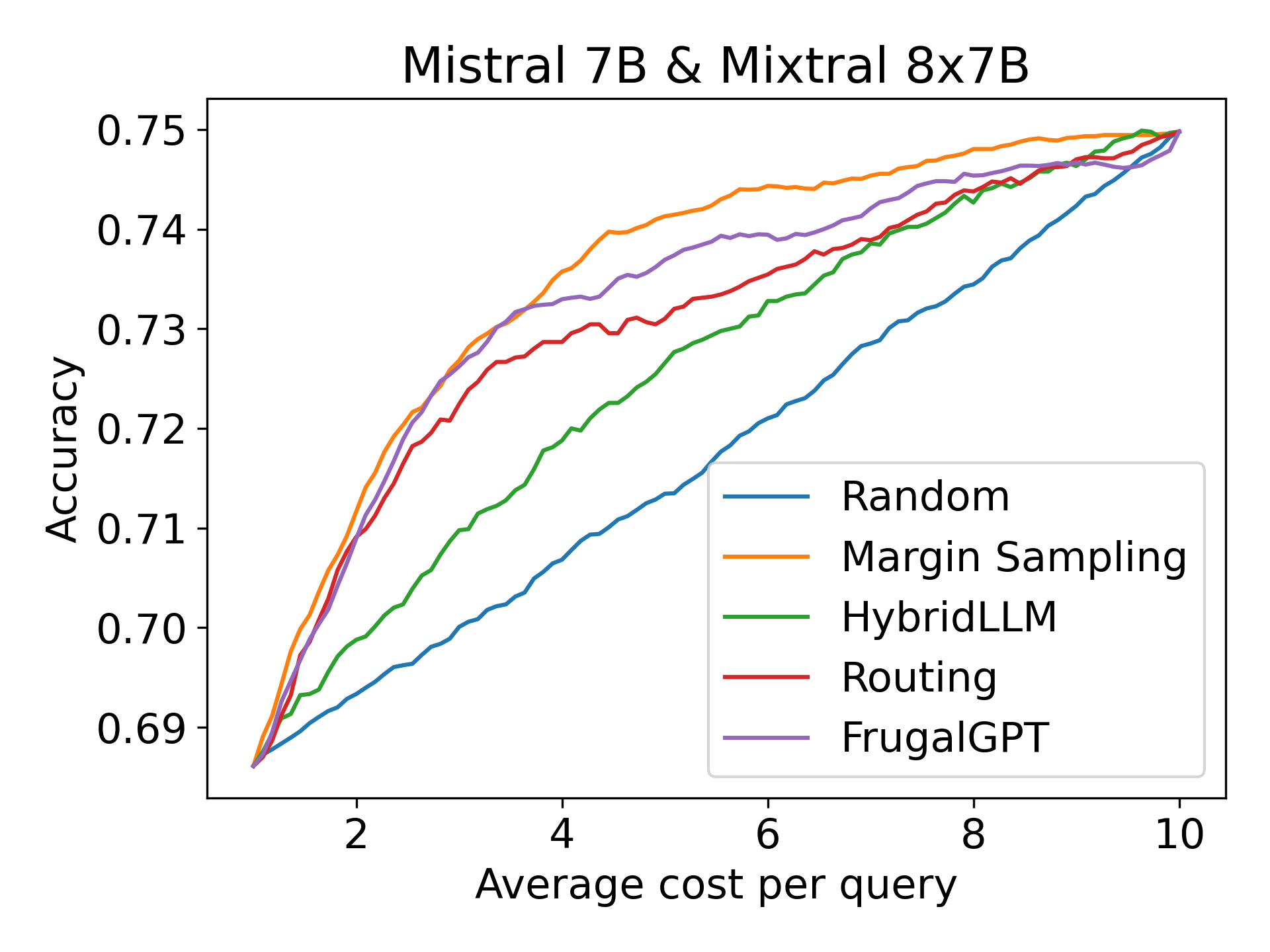}}
{\includegraphics[height=140pt]{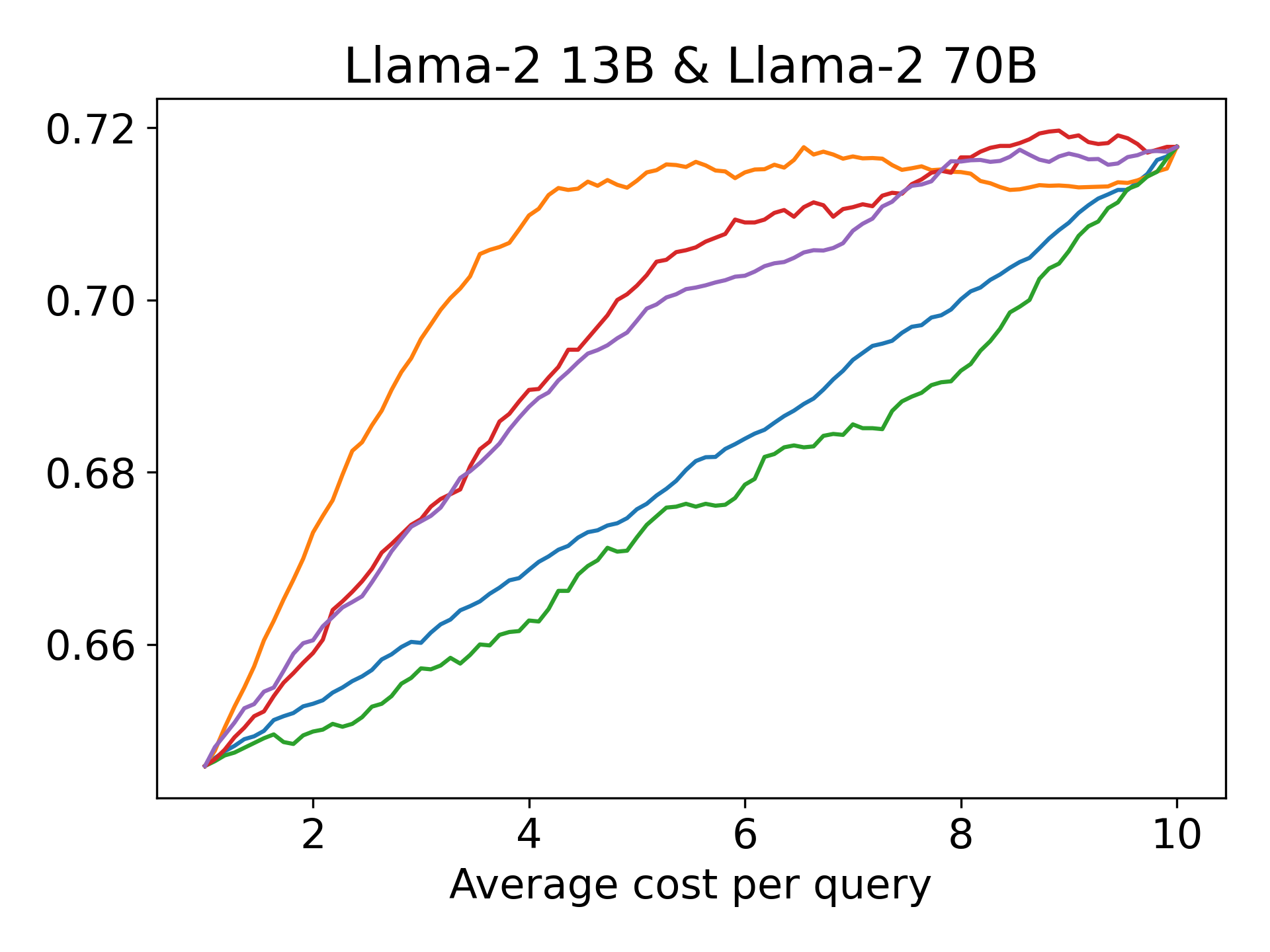}}
{\includegraphics[height=140pt]{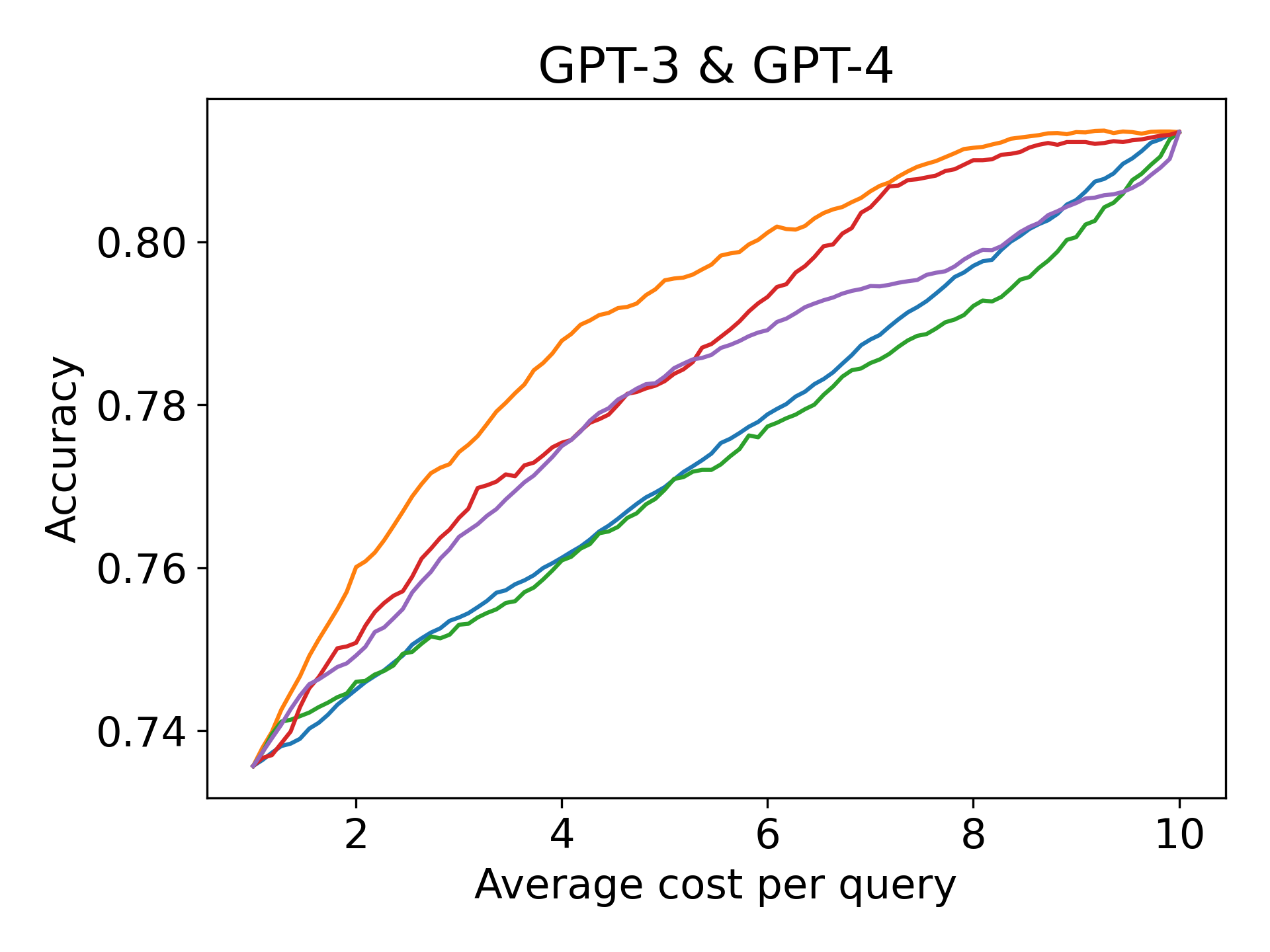}} 

    \caption{Accuracy curve with respect to budgets, in the multi-task setting.}
    \label{plots_multitask}
\end{figure*}

Figure~\ref{plots_multitask} and Table~\ref{multitask} show the results for our multi-task experiments. We see again that Margin Sampling has the best performance. This shows the versatility of this method, that it can be applied across tasks with ease. In contrast, HybridLLM has a poor performance both for Llama-2 and OpenAI models. We found these results still hold when using 5,000 data-points as training data (Appendix~\ref{multiple-long}).

\begin{table}[]
\centering
\begin{tabular}{lccc}
\toprule
\multicolumn{1}{c}{} & \textbf{Mistral} & \textbf{Llama-2} & \textbf{OpenAI} \\
\midrule
Random               & 0.718            & 0.681            & 0.775           \\
Router               & 0.731            & {\ul 0.696}      & {\ul 0.786}     \\
HybridLLM            & 0.726            & 0.676            & 0.773           \\
FrugalGPT            & {\ul 0.733}      & 0.694            & 0.781           \\
Margin Sampling      & \textbf{0.736}   & \textbf{0.704}   & \textbf{0.792} \\
\bottomrule
\end{tabular}
\caption{Accuracy (AUC) in the multi-task setting. Methods Router, HybridLLM and FrugalGPT have been trained with $n=1,000$ data-points.}
\label{multitask}
\end{table}

\subsection{Robustness of experiments}
\subsubsection{Investigating the effect of training data}
To ensure that auxiliary models are not being unfairly handicapped by a low-data setting, we train them with double the data points as before ($n = 1,000$), in spite of this being a possibly infeasible amount of data for many researchers and practitioners.

Table~\ref{moredata} shows the averaged AUC results across tasks for the three LLM pairs. The additional data generally improves the performance of the auxiliary models, however, Margin Sampling still performs competitively, achieving best performance for two of the three pairs. Table~\ref{moredata} also shows that the gap with FrugalGPT closes and that HybridLLM has again a limited performance on the OpenAI models.

\begin{table}[]
\centering
\begin{tabular}{rccc}
\toprule
\multicolumn{1}{c}{}                & \textbf{Mistral} & \textbf{Llama-2} & \textbf{OpenAI}      \\
\midrule
Random                              & 0.681            & 0.648            & 0.732                \\
Router                              & 0.694            & 0.661            & 0.741                \\
HybridLLM                           & 0.691            & 0.664            & 0.734 \\
FrugalGPT                           & {\ul 0.695}      & \textbf{0.672}   & {\ul 0.743}                \\
Margin Sampling & \textbf{0.697}   & {\ul 0.670}      & \textbf{0.751}    \\
\bottomrule
\end{tabular}
\caption{Accuracy (AUC, averaged across tasks) when Router, HybridLLM and FrugalGPT have been trained with 1,000 data-points.}
\label{moredata}
\end{table}

\subsubsection{Investigated the effect of cost}
While previous studies typically only study a single cost setting~\citep{sakota,frugalgpt,automix,hybrid}, this could influence findings. We carry out further experiments with alternative cost schemes for the three LLM pairs across all tasks.~\footnote{We note that, at the time of writing, the pricing ratio for the OpenAI models that we used (GPT-3 and GPT-4) is $\frac{\$30/\text{million tokens}}{\$2/\text{million tokens}}=15$} The cost ratio $r_{\text{cost}}=c_{\text{l}}/c_{\text{s}}$ is sufficient to parameterise the effect of cost. We need only vary $c_{\text{l}}$, and maintain $c_{\text{s}}=1$. We study values of $c_{\text{l}} = 2$, 5 and 20.

Table~\ref{cost} shows the results of these experiments. Independently of the cost, Margin Sampling appears the best cascading strategy (against FrugalGPT). In addition, we observe that Margin Sampling is the best overall strategy for $r_{\text{cost}} \geq 5$; for $r_{\text{cost}} = 2$, routing strategies could be preferred. Intuitively, cascading needs the cost of the small LLM to be relatively cheap enough to not sacrifice too many calls to the large LLM ($p_c < p_r$).


\begin{table}[]
\centering

\begin{tabular}{rccccccccc}
\toprule
\multicolumn{1}{c}{}                & \multicolumn{3}{c}{\textbf{Mistral}}                   & \multicolumn{3}{c}{\textbf{Llama-2}}             & \multicolumn{3}{c}{\textbf{OpenAI}}                                \\
                                    & \footnotesize $c_{\text{l}}$=2            & \footnotesize $c_{\text{l}}$=5                  & \footnotesize $c_{\text{l}}$=20           &  \footnotesize $c_{\text{l}}$=2            & \footnotesize $c_{\text{l}}$=5            & \footnotesize $c_{\text{l}}$=20           & \footnotesize $c_{\text{l}}$=2                  & \footnotesize $c_{\text{l}}$=5                  & \footnotesize $c_{\text{l}}$=20                 \\
                                    \midrule
Random                              & 0.681          & 0.681                & 0.681          & 0.648          & 0.648          & 0.648          & 0.732                & 0.732                & 0.732                \\
Router                              & \textbf{0.690} & {\ul 0.690}          & 0.689          & {\ul 0.657}    & 0.657          & 0.657          & \textbf{0.737}       & {\ul 0.737}                & 0.736                \\
HybridLLM                           & {\ul 0.689}    & 0.688                & 0.688          & \textbf{0.661} & {\ul 0.661}    & 0.661          & 0.732 & 0.732 & 0.731 \\
FrugalGPT                           & 0.677          & 0.687       & {\ul 0.691}    & 0.650          & 0.660          & {\ul 0.664}    & 0.721                & 0.735                & {\ul 0.740}                \\
\multicolumn{1}{l}{Margin Sampling} & 0.683 & \textbf{0.694} & \textbf{0.698} & 0.654          & \textbf{0.667} & \textbf{0.671} & {\ul 0.734}                & \textbf{0.747}       & \textbf{0.752}    \\
\bottomrule
\end{tabular}
\caption{Accuracy (AUC, averaged across datasets) under cost schemes $c_{\text{s}}=1$ and varying $c_{\text{l}}$.}
\label{cost}
\end{table}
\section{Discussion}

In this paper, we have used a simple measure of confidence of the small LLM, known as Margin Sampling, to estimate the uncertainty of an LLM output in short-generation tasks, and we leave for future work the generalisation of this approach to long-generation tasks. Our experiments show that Margin Sampling performs consistently well on a range of short-generation tasks relevant to researchers and practitioners, such as QA and multiple-choice/classification tasks.

Existing approaches on the optimisation of LLM calls require training an auxiliary model. We hypothesised that these approaches introduce additional complexity that obfuscates their understanding, and our findings that HybridLLM performs poorly with the OpenAI models lends credibility to this hypothesis. In contrast, we propose moving towards research that leverages information within the LLM itself. Classic notions of the uncertainty of the LLM's generation, such as perplexity, margin or entropy, could be relevant signals to help optimise LLM calls. A natural extension of our work is to generalise it for an arbitrary task length, for which it may be that a global notion of uncertainty also heavily depends on the first token. 

It was beyond the scope of this work to investigate cascading and routing strategies of three or more LLMs. However, we hypothesise our approach may still perform well, as Margin Sampling has shown to be robust to different-sized LLMs.

\section{Conclusions}
We have proposed a method for LLM call optimisation that achieves superior performance without the need of an auxiliary model. To the best of our knowledge, the field of LLM call optimisation has not yet gained widespread adoption among practitioners. This may be due to the limited versatility and increased complexity of previous solutions. In contrast, our proposed simple approach can be easily and quickly implemented with most commercial LLMs. We believe that our findings could encourage a new direction in the research of LLM call optimisations. 

\newpage
\newpage

\bibliography{references}

\begin{thebibliography}{35}
\providecommand{\natexlab}[1]{#1}
\providecommand{\url}[1]{\texttt{#1}}
\expandafter\ifx\csname urlstyle\endcsname\relax
  \providecommand{\doi}[1]{doi: #1}\else
  \providecommand{\doi}{doi: \begingroup \urlstyle{rm}\Url}\fi

\bibitem[Baan et~al.(2023)Baan, Daheim, Ilia, Ulmer, Li, Fern{\'{a}}ndez, Plank, Sennrich, Zerva, and Aziz]{uncertainty2}
Joris Baan, Nico Daheim, Evgenia Ilia, Dennis Ulmer, Haau{-}Sing Li, Raquel Fern{\'{a}}ndez, Barbara Plank, Rico Sennrich, Chrysoula Zerva, and Wilker Aziz.
\newblock Uncertainty in natural language generation: From theory to applications.
\newblock \emph{CoRR}, abs/2307.15703, 2023.
\newblock \doi{10.48550/ARXIV.2307.15703}.
\newblock URL \url{https://doi.org/10.48550/arXiv.2307.15703}.

\bibitem[Baykal et~al.(2023)Baykal, Trinh, Iliopoulos, Menghani, and Vee]{baykal_robust_2023}
Cenk Baykal, Khoa Trinh, Fotis Iliopoulos, Gaurav Menghani, and Erik Vee.
\newblock Robust active distillation.
\newblock In \emph{The Eleventh International Conference on Learning Representations, {ICLR} 2023, Kigali, Rwanda, May 1-5, 2023}. OpenReview.net, 2023.
\newblock URL \url{https://openreview.net/pdf?id=ALDM5SN2r7M}.

\bibitem[Brown et~al.(2020)Brown, Mann, Ryder, Subbiah, Kaplan, Dhariwal, Neelakantan, Shyam, Sastry, Askell, Agarwal, Herbert{-}Voss, Krueger, Henighan, Child, Ramesh, Ziegler, Wu, Winter, Hesse, Chen, Sigler, Litwin, Gray, Chess, Clark, Berner, McCandlish, Radford, Sutskever, and Amodei]{gpt-3}
Tom~B. Brown, Benjamin Mann, Nick Ryder, Melanie Subbiah, Jared Kaplan, Prafulla Dhariwal, Arvind Neelakantan, Pranav Shyam, Girish Sastry, Amanda Askell, Sandhini Agarwal, Ariel Herbert{-}Voss, Gretchen Krueger, Tom Henighan, Rewon Child, Aditya Ramesh, Daniel~M. Ziegler, Jeffrey Wu, Clemens Winter, Christopher Hesse, Mark Chen, Eric Sigler, Mateusz Litwin, Scott Gray, Benjamin Chess, Jack Clark, Christopher Berner, Sam McCandlish, Alec Radford, Ilya Sutskever, and Dario Amodei.
\newblock Language models are few-shot learners.
\newblock In Hugo Larochelle, Marc'Aurelio Ranzato, Raia Hadsell, Maria{-}Florina Balcan, and Hsuan{-}Tien Lin (eds.), \emph{Advances in Neural Information Processing Systems 33: Annual Conference on Neural Information Processing Systems 2020, NeurIPS 2020, December 6-12, 2020, virtual}, 2020.
\newblock URL \url{https://proceedings.neurips.cc/paper/2020/hash/1457c0d6bfcb4967418bfb8ac142f64a-Abstract.html}.

\bibitem[Bucila et~al.(2006)Bucila, Caruana, and Niculescu{-}Mizil]{KD1}
Cristian Bucila, Rich Caruana, and Alexandru Niculescu{-}Mizil.
\newblock Model compression.
\newblock In Tina Eliassi{-}Rad, Lyle~H. Ungar, Mark Craven, and Dimitrios Gunopulos (eds.), \emph{Proceedings of the Twelfth {ACM} {SIGKDD} International Conference on Knowledge Discovery and Data Mining, Philadelphia, PA, USA, August 20-23, 2006}, pp.\  535--541. {ACM}, 2006.
\newblock \doi{10.1145/1150402.1150464}.
\newblock URL \url{https://doi.org/10.1145/1150402.1150464}.

\bibitem[Chen et~al.(2023)Chen, Zaharia, and Zou]{frugalgpt}
Lingjiao Chen, Matei Zaharia, and James Zou.
\newblock Frugalgpt: How to use large language models while reducing cost and improving performance.
\newblock \emph{CoRR}, abs/2305.05176, 2023.
\newblock \doi{10.48550/ARXIV.2305.05176}.
\newblock URL \url{https://doi.org/10.48550/arXiv.2305.05176}.

\bibitem[Ding et~al.(2024)Ding, Mallick, Wang, Sim, Mukherjee, R{\"u}hle, Lakshmanan, and Awadallah]{hybrid}
Dujian Ding, Ankur Mallick, Chi Wang, Robert Sim, Subhabrata Mukherjee, Victor R{\"u}hle, Laks V.~S. Lakshmanan, and Ahmed~Hassan Awadallah.
\newblock Hybrid {LLM}: Cost-efficient and quality-aware query routing.
\newblock In \emph{The Twelfth International Conference on Learning Representations}, 2024.
\newblock URL \url{https://openreview.net/forum?id=02f3mUtqnM}.

\bibitem[Goodrich et~al.(2019)Goodrich, Rao, Liu, and Saleh]{wikifact}
Ben Goodrich, Vinay Rao, Peter~J. Liu, and Mohammad Saleh.
\newblock Assessing the factual accuracy of generated text.
\newblock In \emph{Proceedings of the 25th ACM SIGKDD International Conference on Knowledge Discovery \& Data Mining}, KDD '19, pp.\  166–175, New York, NY, USA, 2019. Association for Computing Machinery.
\newblock ISBN 9781450362016.
\newblock \doi{10.1145/3292500.3330955}.
\newblock URL \url{https://doi.org/10.1145/3292500.3330955}.

\bibitem[Hinton et~al.(2015)Hinton, Vinyals, and Dean]{KD2}
Geoffrey Hinton, Oriol Vinyals, and Jeffrey Dean.
\newblock Distilling the knowledge in a neural network.
\newblock In \emph{NIPS Deep Learning and Representation Learning Workshop}, 2015.
\newblock URL \url{http://arxiv.org/abs/1503.02531}.

\bibitem[Huang et~al.(2023)Huang, Song, Wang, Chen, and Ma]{uncertainty}
Yuheng Huang, Jiayang Song, Zhijie Wang, Huaming Chen, and Lei Ma.
\newblock Look before you leap: An exploratory study of uncertainty measurement for large language models.
\newblock \emph{CoRR}, abs/2307.10236, 2023.
\newblock \doi{10.48550/ARXIV.2307.10236}.
\newblock URL \url{https://doi.org/10.48550/arXiv.2307.10236}.

\bibitem[Jacob et~al.(2018)Jacob, Kligys, Chen, Zhu, Tang, Howard, Adam, and Kalenichenko]{quantisation}
Benoit Jacob, Skirmantas Kligys, Bo~Chen, Menglong Zhu, Matthew Tang, Andrew~G. Howard, Hartwig Adam, and Dmitry Kalenichenko.
\newblock Quantization and training of neural networks for efficient integer-arithmetic-only inference.
\newblock In \emph{2018 {IEEE} Conference on Computer Vision and Pattern Recognition, {CVPR} 2018, Salt Lake City, UT, USA, June 18-22, 2018}, pp.\  2704--2713. Computer Vision Foundation / {IEEE} Computer Society, 2018.
\newblock \doi{10.1109/CVPR.2018.00286}.
\newblock URL \url{http://openaccess.thecvf.com/content\_cvpr\_2018/html/Jacob\_Quantization\_and\_Training\_CVPR\_2018\_paper.html}.

\bibitem[Jiang et~al.(2023)Jiang, Sablayrolles, Mensch, Bamford, Chaplot, de~Las~Casas, Bressand, Lengyel, Lample, Saulnier, Lavaud, Lachaux, Stock, Scao, Lavril, Wang, Lacroix, and Sayed]{mistral}
Albert~Q. Jiang, Alexandre Sablayrolles, Arthur Mensch, Chris Bamford, Devendra~Singh Chaplot, Diego de~Las~Casas, Florian Bressand, Gianna Lengyel, Guillaume Lample, Lucile Saulnier, L{\'{e}}lio~Renard Lavaud, Marie{-}Anne Lachaux, Pierre Stock, Teven~Le Scao, Thibaut Lavril, Thomas Wang, Timoth{\'{e}}e Lacroix, and William~El Sayed.
\newblock Mistral 7b.
\newblock \emph{CoRR}, abs/2310.06825, 2023.
\newblock \doi{10.48550/ARXIV.2310.06825}.
\newblock URL \url{https://doi.org/10.48550/arXiv.2310.06825}.

\bibitem[Jiang et~al.(2024)Jiang, Sablayrolles, Roux, Mensch, Savary, Bamford, Chaplot, de~Las~Casas, Hanna, Bressand, Lengyel, Bour, Lample, Lavaud, Saulnier, Lachaux, Stock, Subramanian, Yang, Antoniak, Scao, Gervet, Lavril, Wang, Lacroix, and Sayed]{mixtral}
Albert~Q. Jiang, Alexandre Sablayrolles, Antoine Roux, Arthur Mensch, Blanche Savary, Chris Bamford, Devendra~Singh Chaplot, Diego de~Las~Casas, Emma~Bou Hanna, Florian Bressand, Gianna Lengyel, Guillaume Bour, Guillaume Lample, L{\'{e}}lio~Renard Lavaud, Lucile Saulnier, Marie{-}Anne Lachaux, Pierre Stock, Sandeep Subramanian, Sophia Yang, Szymon Antoniak, Teven~Le Scao, Th{\'{e}}ophile Gervet, Thibaut Lavril, Thomas Wang, Timoth{\'{e}}e Lacroix, and William~El Sayed.
\newblock Mixtral of experts.
\newblock \emph{CoRR}, abs/2401.04088, 2024.
\newblock \doi{10.48550/ARXIV.2401.04088}.
\newblock URL \url{https://doi.org/10.48550/arXiv.2401.04088}.

\bibitem[Kwiatkowski et~al.(2019)Kwiatkowski, Palomaki, Redfield, Collins, Parikh, Alberti, Epstein, Polosukhin, Devlin, Lee, Toutanova, Jones, Kelcey, Chang, Dai, Uszkoreit, Le, and Petrov]{DBLP:journals/tacl/KwiatkowskiPRCP19}
Tom Kwiatkowski, Jennimaria Palomaki, Olivia Redfield, Michael Collins, Ankur~P. Parikh, Chris Alberti, Danielle Epstein, Illia Polosukhin, Jacob Devlin, Kenton Lee, Kristina Toutanova, Llion Jones, Matthew Kelcey, Ming{-}Wei Chang, Andrew~M. Dai, Jakob Uszkoreit, Quoc Le, and Slav Petrov.
\newblock Natural questions: a benchmark for question answering research.
\newblock \emph{Trans. Assoc. Comput. Linguistics}, 7:\penalty0 452--466, 2019.
\newblock \doi{10.1162/TACL\_A\_00276}.
\newblock URL \url{https://doi.org/10.1162/tacl\_a\_00276}.

\bibitem[Leviathan et~al.(2023)Leviathan, Kalman, and Matias]{speculative}
Yaniv Leviathan, Matan Kalman, and Yossi Matias.
\newblock Fast inference from transformers via speculative decoding.
\newblock In Andreas Krause, Emma Brunskill, Kyunghyun Cho, Barbara Engelhardt, Sivan Sabato, and Jonathan Scarlett (eds.), \emph{International Conference on Machine Learning, {ICML} 2023, 23-29 July 2023, Honolulu, Hawaii, {USA}}, volume 202 of \emph{Proceedings of Machine Learning Research}, pp.\  19274--19286. {PMLR}, 2023.
\newblock URL \url{https://proceedings.mlr.press/v202/leviathan23a.html}.

\bibitem[Liang et~al.(2022)Liang, Bommasani, Lee, Tsipras, Soylu, Yasunaga, Zhang, Narayanan, Wu, Kumar, Newman, Yuan, Yan, Zhang, Cosgrove, Manning, R{\'{e}}, Acosta{-}Navas, Hudson, Zelikman, Durmus, Ladhak, Rong, Ren, Yao, Wang, Santhanam, Orr, Zheng, Y{\"{u}}ksekg{\"{o}}n{\"{u}}l, Suzgun, Kim, Guha, Chatterji, Khattab, Henderson, Huang, Chi, Xie, Santurkar, Ganguli, Hashimoto, Icard, Zhang, Chaudhary, Wang, Li, Mai, Zhang, and Koreeda]{helm}
Percy Liang, Rishi Bommasani, Tony Lee, Dimitris Tsipras, Dilara Soylu, Michihiro Yasunaga, Yian Zhang, Deepak Narayanan, Yuhuai Wu, Ananya Kumar, Benjamin Newman, Binhang Yuan, Bobby Yan, Ce~Zhang, Christian Cosgrove, Christopher~D. Manning, Christopher R{\'{e}}, Diana Acosta{-}Navas, Drew~A. Hudson, Eric Zelikman, Esin Durmus, Faisal Ladhak, Frieda Rong, Hongyu Ren, Huaxiu Yao, Jue Wang, Keshav Santhanam, Laurel~J. Orr, Lucia Zheng, Mert Y{\"{u}}ksekg{\"{o}}n{\"{u}}l, Mirac Suzgun, Nathan Kim, Neel Guha, Niladri~S. Chatterji, Omar Khattab, Peter Henderson, Qian Huang, Ryan Chi, Sang~Michael Xie, Shibani Santurkar, Surya Ganguli, Tatsunori Hashimoto, Thomas Icard, Tianyi Zhang, Vishrav Chaudhary, William Wang, Xuechen Li, Yifan Mai, Yuhui Zhang, and Yuta Koreeda.
\newblock Holistic evaluation of language models.
\newblock \emph{CoRR}, abs/2211.09110, 2022.
\newblock \doi{10.48550/ARXIV.2211.09110}.
\newblock URL \url{https://doi.org/10.48550/arXiv.2211.09110}.

\bibitem[Lu et~al.(2023)Lu, Yuan, Lin, Lin, Yuan, Zhou, and Zhou]{routingexpert}
Keming Lu, Hongyi Yuan, Runji Lin, Junyang Lin, Zheng Yuan, Chang Zhou, and Jingren Zhou.
\newblock Routing to the expert: Efficient reward-guided ensemble of large language models.
\newblock \emph{CoRR}, abs/2311.08692, 2023.
\newblock \doi{10.48550/ARXIV.2311.08692}.
\newblock URL \url{https://doi.org/10.48550/arXiv.2311.08692}.

\bibitem[Luo et~al.(2004)Luo, Kramer, Samson, Remsen, Goldgof, Hall, and Hopkins]{luo_active_2004}
Tong Luo, K.~Kramer, S.~Samson, A.~Remsen, D.B. Goldgof, L.O. Hall, and T.~Hopkins.
\newblock Active learning to recognize multiple types of plankton.
\newblock In \emph{Proceedings of the 17th {International} {Conference} on {Pattern} {Recognition}, 2004. {ICPR} 2004.}, volume~3, pp.\  478--481 Vol.3, August 2004.
\newblock \doi{10.1109/ICPR.2004.1334570}.
\newblock ISSN: 1051-4651.

\bibitem[Madaan et~al.(2023)Madaan, Aggarwal, Anand, Potharaju, Mishra, Zhou, Gupta, Rajagopal, Kappaganthu, Yang, Upadhyay, Mausam, and Faruqui]{automix}
Aman Madaan, Pranjal Aggarwal, Ankit Anand, Srividya~Pranavi Potharaju, Swaroop Mishra, Pei Zhou, Aditya Gupta, Dheeraj Rajagopal, Karthik Kappaganthu, Yiming Yang, Shyam Upadhyay, Mausam, and Manaal Faruqui.
\newblock Automix: Automatically mixing language models.
\newblock \emph{CoRR}, abs/2310.12963, 2023.
\newblock \doi{10.48550/ARXIV.2310.12963}.
\newblock URL \url{https://doi.org/10.48550/arXiv.2310.12963}.

\bibitem[Mihaylov et~al.(2018)Mihaylov, Clark, Khot, and Sabharwal]{openbook}
Todor Mihaylov, Peter Clark, Tushar Khot, and Ashish Sabharwal.
\newblock Can a suit of armor conduct electricity? {A} new dataset for open book question answering.
\newblock In Ellen Riloff, David Chiang, Julia Hockenmaier, and Jun'ichi Tsujii (eds.), \emph{Proceedings of the 2018 Conference on Empirical Methods in Natural Language Processing, Brussels, Belgium, October 31 - November 4, 2018}, pp.\  2381--2391. Association for Computational Linguistics, 2018.
\newblock \doi{10.18653/v1/d18-1260}.
\newblock URL \url{https://doi.org/10.18653/v1/d18-1260}.

\bibitem[Ni et~al.(2019)Ni, Li, and McAuley]{cr}
Jianmo Ni, Jiacheng Li, and Julian McAuley.
\newblock Justifying recommendations using distantly-labeled reviews and fine-grained aspects.
\newblock In Kentaro Inui, Jing Jiang, Vincent Ng, and Xiaojun Wan (eds.), \emph{Proceedings of the 2019 Conference on Empirical Methods in Natural Language Processing and the 9th International Joint Conference on Natural Language Processing (EMNLP-IJCNLP)}, pp.\  188--197, Hong Kong, China, November 2019. Association for Computational Linguistics.
\newblock \doi{10.18653/v1/D19-1018}.
\newblock URL \url{https://aclanthology.org/D19-1018}.

\bibitem[OpenAI(2023)]{gpt-4}
OpenAI.
\newblock {GPT-4} technical report.
\newblock \emph{CoRR}, abs/2303.08774, 2023.
\newblock \doi{10.48550/ARXIV.2303.08774}.
\newblock URL \url{https://doi.org/10.48550/arXiv.2303.08774}.

\bibitem[Pang \& Lee(2005)Pang and Lee]{pang_seeing_2005}
Bo~Pang and Lillian Lee.
\newblock Seeing {Stars}: {Exploiting} {Class} {Relationships} for {Sentiment} {Categorization} with {Respect} to {Rating} {Scales}.
\newblock In \emph{Proceedings of the 43rd {Annual} {Meeting} of the {Association} for {Computational} {Linguistics} ({ACL}'05)}, pp.\  115--124, Ann Arbor, Michigan, June 2005. Association for Computational Linguistics.
\newblock \doi{10.3115/1219840.1219855}.
\newblock URL \url{https://aclanthology.org/P05-1015}.

\bibitem[Petroni et~al.(2019)Petroni, Rockt{\"{a}}schel, Riedel, Lewis, Bakhtin, Wu, and Miller]{wikifact2}
Fabio Petroni, Tim Rockt{\"{a}}schel, Sebastian Riedel, Patrick S.~H. Lewis, Anton Bakhtin, Yuxiang Wu, and Alexander~H. Miller.
\newblock Language models as knowledge bases?
\newblock In Kentaro Inui, Jing Jiang, Vincent Ng, and Xiaojun Wan (eds.), \emph{Proceedings of the 2019 Conference on Empirical Methods in Natural Language Processing and the 9th International Joint Conference on Natural Language Processing, {EMNLP-IJCNLP} 2019, Hong Kong, China, November 3-7, 2019}, pp.\  2463--2473. Association for Computational Linguistics, 2019.
\newblock \doi{10.18653/V1/D19-1250}.
\newblock URL \url{https://doi.org/10.18653/v1/D19-1250}.

\bibitem[Ram{\'{\i}}rez et~al.(2023)Ram{\'{\i}}rez, Lindemann, Birch, and Titov]{cache-distil}
Guillem Ram{\'{\i}}rez, Matthias Lindemann, Alexandra Birch, and Ivan Titov.
\newblock Cache {\&} distil: Optimising {API} calls to large language models.
\newblock \emph{CoRR}, abs/2310.13561, 2023.
\newblock \doi{10.48550/ARXIV.2310.13561}.
\newblock URL \url{https://doi.org/10.48550/arXiv.2310.13561}.

\bibitem[Sakota et~al.(2023)Sakota, Peyrard, and West]{sakota}
Marija Sakota, Maxime Peyrard, and Robert West.
\newblock Fly-swat or cannon? cost-effective language model choice via meta-modeling.
\newblock \emph{CoRR}, abs/2308.06077, 2023.
\newblock \doi{10.48550/ARXIV.2308.06077}.
\newblock URL \url{https://doi.org/10.48550/arXiv.2308.06077}.

\bibitem[Sanh et~al.(2019)Sanh, Debut, Chaumond, and Wolf]{distilbert}
Victor Sanh, Lysandre Debut, Julien Chaumond, and Thomas Wolf.
\newblock Distilbert, a distilled version of {BERT:} smaller, faster, cheaper and lighter.
\newblock \emph{CoRR}, abs/1910.01108, 2019.
\newblock URL \url{http://arxiv.org/abs/1910.01108}.

\bibitem[Scheffer et~al.(2001)Scheffer, Decomain, and Wrobel]{scheffer_active_2001}
Tobias Scheffer, Christian Decomain, and Stefan Wrobel.
\newblock Active {Hidden} {Markov} {Models} for {Information} {Extraction}.
\newblock In Frank Hoffmann, David~J. Hand, Niall Adams, Douglas Fisher, and Gabriela Guimaraes (eds.), \emph{Advances in {Intelligent} {Data} {Analysis}}, Lecture {Notes} in {Computer} {Science}, pp.\  309--318, Berlin, Heidelberg, 2001. Springer.
\newblock ISBN 978-3-540-44816-7.
\newblock \doi{10.1007/3-540-44816-0_31}.

\bibitem[Shao et~al.(2015)Shao, Doucet, and Caruso]{shao_universality_2015}
Bo~Shao, Lorna Doucet, and David~R. Caruso.
\newblock Universality {Versus} {Cultural} {Specificity} of {Three} {Emotion} {Domains}: {Some} {Evidence} {Based} on the {Cascading} {Model} of {Emotional} {Intelligence}.
\newblock \emph{Journal of Cross-Cultural Psychology}, 46\penalty0 (2):\penalty0 229--251, February 2015.
\newblock ISSN 0022-0221.
\newblock \doi{10.1177/0022022114557479}.
\newblock URL \url{https://doi.org/10.1177/0022022114557479}.
\newblock Publisher: SAGE Publications Inc.

\bibitem[Socher et~al.(2013)Socher, Perelygin, Wu, Chuang, Manning, Ng, and Potts]{sst2}
Richard Socher, Alex Perelygin, Jean Wu, Jason Chuang, Christopher~D. Manning, Andrew~Y. Ng, and Christopher Potts.
\newblock Recursive deep models for semantic compositionality over a sentiment treebank.
\newblock In \emph{Proceedings of the 2013 Conference on Empirical Methods in Natural Language Processing, {EMNLP} 2013, 18-21 October 2013, Grand Hyatt Seattle, Seattle, Washington, USA, {A} meeting of SIGDAT, a Special Interest Group of the {ACL}}, pp.\  1631--1642. {ACL}, 2013.
\newblock URL \url{https://aclanthology.org/D13-1170/}.

\bibitem[Thorne et~al.(2018)Thorne, Vlachos, Christodoulopoulos, and Mittal]{thorne_fever_2018}
James Thorne, Andreas Vlachos, Christos Christodoulopoulos, and Arpit Mittal.
\newblock {FEVER:} a large-scale dataset for fact extraction and verification.
\newblock In Marilyn~A. Walker, Heng Ji, and Amanda Stent (eds.), \emph{Proceedings of the 2018 Conference of the North American Chapter of the Association for Computational Linguistics: Human Language Technologies, {NAACL-HLT} 2018, New Orleans, Louisiana, USA, June 1-6, 2018, Volume 1 (Long Papers)}, pp.\  809--819. Association for Computational Linguistics, 2018.
\newblock \doi{10.18653/v1/n18-1074}.
\newblock URL \url{https://doi.org/10.18653/v1/n18-1074}.

\bibitem[Touvron et~al.(2023)Touvron, Martin, Stone, Albert, Almahairi, Babaei, Bashlykov, Batra, Bhargava, Bhosale, Bikel, Blecher, Canton{-}Ferrer, Chen, Cucurull, Esiobu, Fernandes, Fu, Fu, Fuller, Gao, Goswami, Goyal, Hartshorn, Hosseini, Hou, Inan, Kardas, Kerkez, Khabsa, Kloumann, Korenev, Koura, Lachaux, Lavril, Lee, Liskovich, Lu, Mao, Martinet, Mihaylov, Mishra, Molybog, Nie, Poulton, Reizenstein, Rungta, Saladi, Schelten, Silva, Smith, Subramanian, Tan, Tang, Taylor, Williams, Kuan, Xu, Yan, Zarov, Zhang, Fan, Kambadur, Narang, Rodriguez, Stojnic, Edunov, and Scialom]{llama2}
Hugo Touvron, Louis Martin, Kevin Stone, Peter Albert, Amjad Almahairi, Yasmine Babaei, Nikolay Bashlykov, Soumya Batra, Prajjwal Bhargava, Shruti Bhosale, Dan Bikel, Lukas Blecher, Cristian Canton{-}Ferrer, Moya Chen, Guillem Cucurull, David Esiobu, Jude Fernandes, Jeremy Fu, Wenyin Fu, Brian Fuller, Cynthia Gao, Vedanuj Goswami, Naman Goyal, Anthony Hartshorn, Saghar Hosseini, Rui Hou, Hakan Inan, Marcin Kardas, Viktor Kerkez, Madian Khabsa, Isabel Kloumann, Artem Korenev, Punit~Singh Koura, Marie{-}Anne Lachaux, Thibaut Lavril, Jenya Lee, Diana Liskovich, Yinghai Lu, Yuning Mao, Xavier Martinet, Todor Mihaylov, Pushkar Mishra, Igor Molybog, Yixin Nie, Andrew Poulton, Jeremy Reizenstein, Rashi Rungta, Kalyan Saladi, Alan Schelten, Ruan Silva, Eric~Michael Smith, Ranjan Subramanian, Xiaoqing~Ellen Tan, Binh Tang, Ross Taylor, Adina Williams, Jian~Xiang Kuan, Puxin Xu, Zheng Yan, Iliyan Zarov, Yuchen Zhang, Angela Fan, Melanie Kambadur, Sharan Narang, Aur{\'{e}}lien Rodriguez, Robert Stojnic, Sergey Edunov,
  and Thomas Scialom.
\newblock Llama 2: Open foundation and fine-tuned chat models.
\newblock \emph{CoRR}, abs/2307.09288, 2023.
\newblock \doi{10.48550/ARXIV.2307.09288}.
\newblock URL \url{https://doi.org/10.48550/arXiv.2307.09288}.

\bibitem[Wang et~al.(2024)Wang, Zhang, Sui, Tu, Liu, and Kang]{survey}
Can Wang, Bolin Zhang, Dianbo Sui, Zhiying Tu, Xiaoyu Liu, and Jiabao Kang.
\newblock A survey on effective invocation methods of massive {LLM} services.
\newblock \emph{CoRR}, abs/2402.03408, 2024.
\newblock \doi{10.48550/ARXIV.2402.03408}.
\newblock URL \url{https://doi.org/10.48550/arXiv.2402.03408}.

\bibitem[Weston et~al.(2016)Weston, Bordes, Chopra, and Mikolov]{babi}
Jason Weston, Antoine Bordes, Sumit Chopra, and Tom{\'{a}}s Mikolov.
\newblock Towards ai-complete question answering: {A} set of prerequisite toy tasks.
\newblock In Yoshua Bengio and Yann LeCun (eds.), \emph{4th International Conference on Learning Representations, {ICLR} 2016, San Juan, Puerto Rico, May 2-4, 2016, Conference Track Proceedings}, 2016.
\newblock URL \url{http://arxiv.org/abs/1502.05698}.

\bibitem[Yue et~al.(2024)Yue, Zhao, Zhang, Du, and Yao]{qbc}
Murong Yue, Jie Zhao, Min Zhang, Liang Du, and Ziyu Yao.
\newblock Large language model cascades with mixture of thought representations for cost-efficient reasoning.
\newblock In \emph{The Twelfth International Conference on Learning Representations}, 2024.
\newblock URL \url{https://openreview.net/forum?id=6okaSfANzh}.

\bibitem[Zhang et~al.(2023)Zhang, Krishna, Awadallah, and Wang]{ecoassistant}
Jieyu Zhang, Ranjay Krishna, Ahmed~Hassan Awadallah, and Chi Wang.
\newblock Ecoassistant: Using {LLM} assistant more affordably and accurately.
\newblock \emph{CoRR}, abs/2310.03046, 2023.
\newblock \doi{10.48550/ARXIV.2310.03046}.
\newblock URL \url{https://doi.org/10.48550/arXiv.2310.03046}.

\end{thebibliography}
\bibliographystyle{colm2024_conference}
\begin{appendix}

\section{Implementation details}
\label{implement}
For methods Routing, HybridLLM and FrugalGPT we use Huggingface's \texttt{AutoModelForSequenceClassification} and the model \texttt{distilbert/distilbert-base-uncased}. For methods Routing, HybridLLM we set the target number of classes to two; for FrugalGPT, we set it to either the number of target classes for classification/multiple-choice problems or two for QA problems. We perform a hyperparameter search (grid search) on a validation set of 500 examples of Openbook and Wikifact, and we find that the different methods have a similar convergence. We decide using learning rate $\mu = 5\times 10^{-4}$, training batch size $m = 16$ and weight decay $\lambda = 0.01$, which is consistent with the reported values of \cite{sakota} and seems to generalise well across tasks. \citet{hybrid}, \citet{frugalgpt} and \citet{routingexpert} do not report hyperparameter values. 
\subsection{Training}
\paragraph{Routing} We generate an answer using the small LLM. Then, we compare it to the gold label. The target class is 1 or 0 depending on the correctness of the answer from the small LLM.
\begin{itemize}
    \item \textbf{Input}: 'Who wrote 'If There Is I Haven't Found It Yet?' 
    \item \textbf{Output}: '0'
    \item Explanation: Llama-2 13B produces an incorrect answer.
\end{itemize}

\paragraph{HybridLLM} We generate an answer using the small and the large LLMs. Then, we use the first token $y_1$ of the gold answer $y$, to obtain the quality gap $H(x) = P_{\text{s}}(y_1 \mid x) - P_{\text{l}}(y_1 \mid x)$. Following \citet{hybrid}, we subtract the median of $H(x)$; then we map it to 1 or 0 depending on $H(x) > H_{\text{median}}$, thus leaving a binary classification problem of uniform classes.

\begin{itemize}
    \item \textbf{Input}: 'Who wrote 'If There Is I Haven't Found It Yet?' 
    \item \textbf{Output}: '0'
    \item Explanation: Llama-2 13B is less likely than Llama-2 70B to produce the right answer.
\end{itemize}

\paragraph{FrugalGPT (classification and multiple-choice)} We train DistilBERT with gold data. During inference, we generate an answer with the small LLM. The score is the probability DistilBERT associates to this class. 

\paragraph{FrugalGPT (QA)} We train a binary classifier that predicts if an answer is correct. To do so, we use as the positive class the gold labels. We generate answers with Llama-2 13B and tag them as either positive or negative class depending on whether they match the gold labels. 

\begin{itemize}
    \item \textbf{Input}: 'Who wrote 'If There Is I Haven't Found It Yet? ANSWER: Anna Funder' 
    \item \textbf{Output}: '0'
    \item Explanation: Anna Funder is an incorrect answer.
\end{itemize}

\section{Datasets}
\label{appendix:datasets}
Table~\ref{datasets} contains some statistics on the datasets used. All the classification datasets are uniformly distributed. For our experiments, we have reserved 1,000 datapoints for training the scorers (Router, HybridLLM and FrugalGPT) and used the remaining of the datasets for online inference. 

\addtolength{\tabcolsep}{-0.1em}
\begin{table*}[]
\begin{tabular}{lccccccccc}
\toprule
\multicolumn{1}{c}{} & \footnotesize \textbf{ISEAR} & \footnotesize \textbf{RT-Polarity} & \footnotesize \textbf{FEVER} & \footnotesize \textbf{CR} & \footnotesize \textbf{SST-2} & \footnotesize \textbf{Openbook} & \footnotesize \textbf{Wikifact} & \footnotesize \textbf{bAbI} & \footnotesize \textbf{NaturalQ} \\
\midrule
\small Total datapoints                        & \small 6132           & \small 8529                 & \small 5289        & \small 3393           & \small 7000           & \small 5457                & \small 5733                      & \small 3000              & \small 2876          \\
\small Number of classes                       & \small 7              & \small 2                    & \small 2           & \small 2              & \small 2              & \small 4                   & \small QA                        & \small QA                & \small QA            \\
\bottomrule
\end{tabular}
\caption{Datasets used.} 
\label{datasets}
\end{table*}

\begin{table*}[]
\centering
\begin{tabular}{lccccccccc}
\toprule
\multicolumn{1}{c}{} & \footnotesize \textbf{ISEAR} & \footnotesize \textbf{RT-Polarity} & \footnotesize \textbf{FEVER} & \footnotesize \textbf{CR} & \footnotesize \textbf{SST-2} & \footnotesize \textbf{Openbook} & \footnotesize \textbf{Wikifact} & \footnotesize \textbf{bAbI} & \footnotesize \textbf{NaturalQ} \\
\midrule
\footnotesize Mistral 7B & \footnotesize	0.557	 & \footnotesize 0.862  & \footnotesize	0.770	 & \footnotesize 0.911	 & \footnotesize 0.854	 & \footnotesize 0.813	 & \footnotesize 0.359	 & \footnotesize 0.560	 & \footnotesize 0.125 \\
\footnotesize Mixtral 8x7B	& \footnotesize  0.655	& \footnotesize 0.889	& \footnotesize 0.779	& \footnotesize 0.936	& \footnotesize 0.906 & \footnotesize 	0.875	& \footnotesize 0.530	& \footnotesize 0.634 & \footnotesize 	0.260 \\
\midrule 
\footnotesize Llama-13b	& \footnotesize 0.599	& \footnotesize 0.798	& \footnotesize 0.613	& \footnotesize 0.902	& \footnotesize 0.867	& \footnotesize 0.556	& \footnotesize 0.416	& \footnotesize 0.525	& \footnotesize 0.212 \\
\footnotesize Llama-70b	& \footnotesize 0.661	& \footnotesize 0.820	& \footnotesize 0.691	& \footnotesize 0.871	& \footnotesize 0.882	& \footnotesize 0.681	& \footnotesize 0.590	& \footnotesize 0.676	& \footnotesize 0.307 \\ \midrule 
\footnotesize GPT-3	& \footnotesize 0.699	& \footnotesize 0.900	& \footnotesize 0.777	& \footnotesize 0.921	& \footnotesize 0.897	& \footnotesize 0.798	& \footnotesize 0.486	& \footnotesize 0.462	& \footnotesize 0.251 \\
\footnotesize GPT-4	& \footnotesize 0.796	& \footnotesize 0.929	& \footnotesize 0.852	& \footnotesize 0.943	& \footnotesize 0.899	& \footnotesize 0.956	& \footnotesize 0.622	& \footnotesize 0.695	& \footnotesize 0.324 \\
\bottomrule
\end{tabular}
\caption{Accuracy of the LLMs in the studied tasks.} 
\label{LLM_datasets}
\end{table*}

\section{LLMs used}
\label{llms_used}
We load the open-source LLMs with Huggingface's \texttt{AutoModelForCausalLM.from\_pretrained}, activating \texttt{load\_in\_4bit}. \\ 
We use models \texttt{meta-llama/Llama-2-13b-hf}, \texttt{meta-llama/Llama-2-70b-hf}, \texttt{mistralai/Mistral-7B-Instruct-v0.2} and \texttt{mistralai/Mixtral-8x7B-Instruct-v0.1}. We set the temperature to 0 and look at the most likely token. 

For ISEAR, RT-Polarity, FEVER, OpenbookQA, SST-2 and CR, we use the prompts from \citet{cache-distil} (0-shot). For Wikifact, NaturalQuestions and bAbI we use the prompts from the HELM benchmark~\citep{helm}.

For the OpenAI models, we use \texttt{davinci-002} (GPT-3) and \texttt{gpt-4}. Annotating all the datasets has a cost of around $\$180$.

Table~\ref{LLM_datasets} contains shows the accuracy of the LLMs across the different tasks.  

\section{Additional results}
\subsection{Multi-task setup}
\label{multiple-long}

We experiment to see how far we can get with Routing, HybridLLM and FrugalGPT with more data. We train these methods with $n=5,000$ datapoints. We find that Margin Sampling still outperforms them in this setup (Table~\ref{multitask_large}). 
\begin{table}[]
\centering
\begin{tabular}{lccc}
\toprule
\multicolumn{1}{c}{} & \textbf{Mistral} & \textbf{Llama-2} & \textbf{OpenAI} \\
\midrule
Random               & 0.718            & 0.682            & 0.777           \\
Router               & {\ul 0.734}            & 0.696      & {\ul 0.790}     \\
HybridLLM            & 0.725            & 0.688            & 0.774          \\
FrugalGPT            & \textbf{0.739}      & \textbf{0.706}            & 0.791           \\
Margin Sampling      & \textbf{0.739}   & {\ul 0.705}   & \textbf{0.794} \\
\bottomrule
\end{tabular}
\caption{Accuracy (AUC) in the multiple-task setting. Methods Router, HybridLLM and FrugalGPT have been trained with $n=5,000$ datapoints.}
\label{multitask_large}
\end{table}

\subsection{Multiple calls to the LLM}
\label{multiple_calls}
We experiment with the method from \citet{qbc}, which estimates the uncertainty of the generation by doing multiple calls to the small LLM. We make 5 calls, sampling with temperature $T=1$. Our results (Table~\ref{qbc}) reveal this method does badly in our setup. The reason for this relies that doing the multiple calls to the small LLM is relatively expensive in our setup with $c_{\text{s}}=1, c_{\text{l}}=10$.

\begin{table*}[]
\centering
\begin{tabular}{lccccccccc}
\toprule
\multicolumn{1}{c}{} & \footnotesize \textbf{ISEAR} & \footnotesize \textbf{RT-Polarity} & \footnotesize \textbf{FEVER} & \footnotesize \textbf{CR} & \footnotesize \textbf{SST-2} & \footnotesize \textbf{Openbook} & \footnotesize \textbf{Wikifact} & \footnotesize \textbf{bAbI} & \footnotesize \textbf{NaturalQ} \\
\midrule
\footnotesize Random	& \footnotesize 0.630	& \footnotesize 0.809		& \footnotesize 0.653	& \footnotesize 	0.885		& \footnotesize 0.873		& \footnotesize 0.617	& \footnotesize 	0.505	& \footnotesize 	0.600		& \footnotesize 0.259 \\
\footnotesize Committee	& \footnotesize 	0.620		& \footnotesize 0.797	& \footnotesize 	0.591		& \footnotesize 0.886		& \footnotesize 0.863		& \footnotesize 0.587		& \footnotesize 0.473		& \footnotesize 0.538	& \footnotesize 	0.234 \\
\bottomrule
\end{tabular}
\caption{Accuracy (AUC) for the Committee method~\citep{qbc}, for Llama-2 13B and Llama-2 70B.}
\label{qbc}
\end{table*}

\end{appendix}

\end{document}